\pdfoutput=1

\documentclass[11pt]{article}

\usepackage[]{acl}

\usepackage{times}
\usepackage{latexsym}

\usepackage[T1]{fontenc}

\usepackage[utf8]{inputenc}

\usepackage{microtype}
\usepackage{arydshln}

\usepackage{inconsolata}

\usepackage[table]{colortbl}

\usepackage{amsmath,amsfonts,bm}

\def\eqref#1{equation~\ref{#1}}

\def\1{\bm{1}}

\DeclareMathAlphabet{\mathsfit}{\encodingdefault}{\sfdefault}{m}{sl}
\SetMathAlphabet{\mathsfit}{bold}{\encodingdefault}{\sfdefault}{bx}{n}

\usepackage{hyperref}
\usepackage{url}
\usepackage{graphicx}
\usepackage{algorithm}
\usepackage{algorithmic}
\usepackage{amsmath}
\usepackage{multirow}
\usepackage{booktabs}
\usepackage{threeparttable}
\usepackage{rotating}
\usepackage{multicol}
\usepackage{makecell}
\usepackage{subcaption}
\usepackage{tikz}
\usepackage{csquotes}
\usepackage{pifont}

\title{DynamicKV: Task-Aware Adaptive KV Cache Compression for\\ Long Context LLMs}

\author{Xiabin Zhou,\space\space Wenbin Wang,\space\space Minyan Zeng,\space\space Jiaxian Guo,\\ \textbf{Xuebo Liu,}\space\space \textbf{Li Shen,}\space\space \textbf{Min Zhang,}\space\space \textbf{Liang Ding}\space\space\\
{\tt\small \{xiabinzhou0625, liangding.liam\}@gmail.com}
}

\newcommand{\itbf}[1]{\textit{\textbf{#1}}}

\begin{document}

\maketitle
\begin{abstract}

Efficient KV cache management in LLMs is crucial for long-context tasks like RAG and summarization. Existing KV cache compression methods enforce a fixed pattern, neglecting task-specific characteristics and reducing the retention of essential information.
However, we observe distinct activation patterns across layers in various tasks, highlighting the need for adaptive strategies tailored to each task's unique demands. Based on this insight, we propose \textbf{DynamicKV}, a method that dynamically optimizes token retention by adjusting the number of tokens retained at each layer to adapt to the specific task. DynamicKV establishes global and per-layer maximum KV cache budgets, temporarily retaining the maximum budget for the current layer, and periodically updating the KV cache sizes of all preceding layers during inference. Our method \textbf{retains only $1.7\%$ of the KV cache size while achieving $\sim90\%$ of the Full KV cache performance} on LongBench. 
Notably, even under extreme compression ($0.9\%$), \textbf{DynamicKV surpasses state-of-the-art (SOTA) methods by 11\% in the Needle-in-a-Haystack test} using Mistral-7B-Instruct-v0.2. The code will be released.
\end{abstract}
\section{Introduction}
Large Language Models (LLMs) \citep{achiam2023gpt} are exerting a considerable influence in the field of natural language processing (NLP), driving advancements in summarization, translation, code generation, etc. \citep{chiang2023vicuna,zhong2023can,peng2023towards,lu2023error,wang2024mathbb}. Recent developments in LLMs \citep{liu2024lost} have been scaled up to handle long contexts, with LlaMA3~\citep{dubey2024llama} processing up to 32K tokens and InternLM~\citep{cai2024internlm2} handling 1M tokens. Scaling LLMs to longer contexts introduces significant latency due to the quadratic complexity of attention. A common solution is to cache key and value (KV) status~\citep{waddington2013kv}, reducing computation. However, this comes at a high memory cost -- for example, caching 100K tokens in LLaMA2-7B~\citep{touvron2023llama2} still requires over 50GB of memory.

\begin{figure*}[!t]
    \centering
    \begin{subfigure}[b]{0.68\textwidth}
        \includegraphics[width=\textwidth]{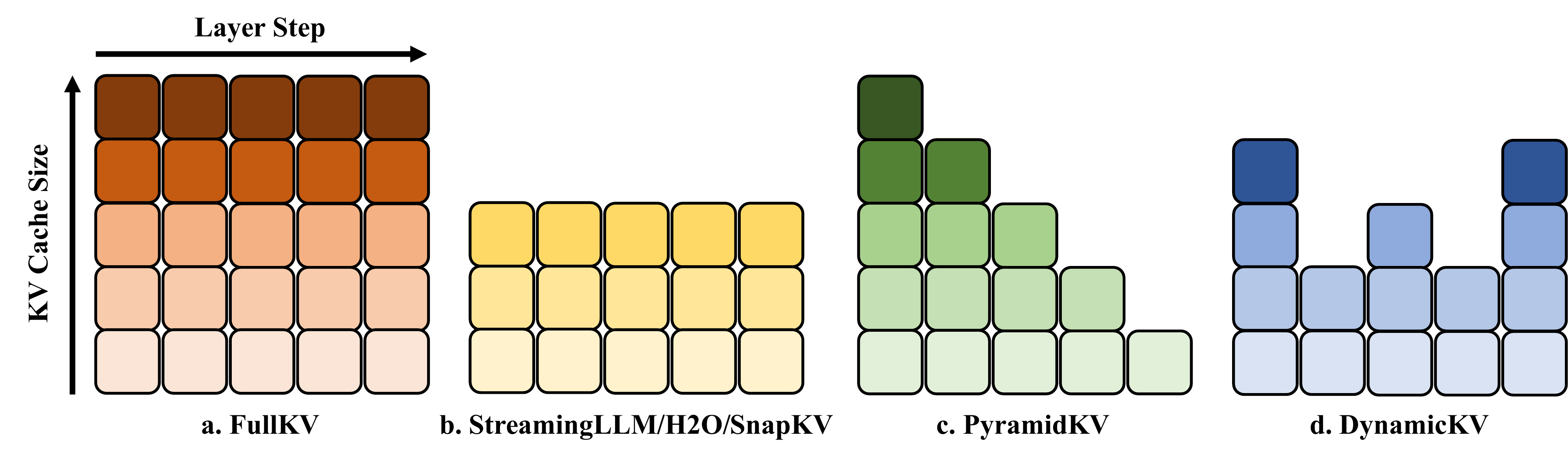} 
        \label{fig:compare_with_exiting_methods_sub1}
    \end{subfigure}
    \hspace{0.05em} 
    \rule[0.5cm]{0.4pt}{2.8cm}
    \hspace{0.05em} 
    \begin{subfigure}[b]{0.26\textwidth}
        \includegraphics[width=\textwidth]{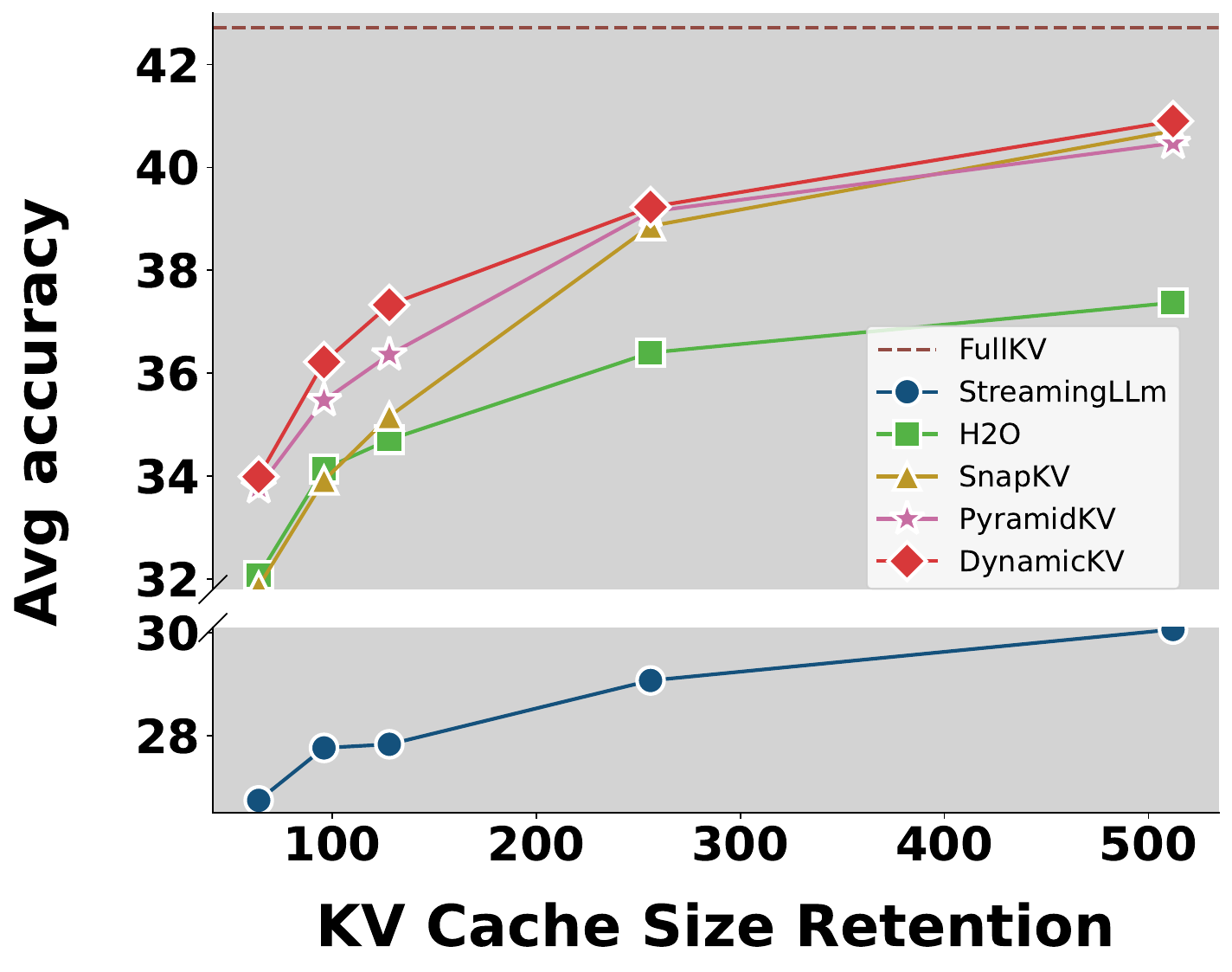}
        \label{fig:compare_with_exiting_methods_sub2}
    \end{subfigure}
    \caption{
        \textbf{Comparison of DynamicKV with traditional methods in maintaining KV cache size across layers.}
        Left: the structure difference:
        (a) Retain all KV cache.
        (b) Fixed KV cache for each layer (e.g., StreamingLLM, H2O, SnapKV).
        (c) Hierarchically decreasing pyramid KV cache retention.
        (d) Ours DynamicKV: layer-aware adaptive KV cache retention.
        Right: average accuracy on different KV cache retention.
    }
    \vspace{-0.3cm}
    \label{figs:compare_with_exiting_methods}
\end{figure*}

To address this issue, recent studies have explored the optimization of KV caching, including KV cache quantization \citep{kang2024gear, hooper2024kvquant}, token dropping \citep{zhang2024h2o, xiao2023efficient}, architectural improvements to Transformers \citep{sun2024you}, KV cache fusion \citep{nawrot2024dynamic}, and hierarchical sharing and constraints\citep{liu2024minicache, brandon2024reducing}. Existing KV cache compression methods enforce a fixed pattern (as shown in Figure~\ref{figs:compare_with_exiting_methods}), such as a hierarchical pyramid structure \citep{zhang2024pyramidkv} or a structure similar to FastGen's fixed internal pattern \citep{ge2023model}, or they fix the length of the KV cache to selectively retain tokens across different layers \citep{zhang2024h2o, li2024snapkv}. However, LLMs require different numbers of layers when handling different types of tasks. For example, for knowledge-based question-answering tasks, only the first few layers are needed to achieve high accuracy, while for complex reasoning tasks (\textit{e.g.,} mathematics and code generation), more layers are often required to achieve higher accuracy~\citep{elhoushi2024layer}. Thus, we raise a question: 
\textit{Do different types of tasks all follow a fixed pattern?}

To examine this question, we aim to systematically investigate the design principles of the KV cache compression across different tasks. Inspired by~\citet{zhang2024pyramidkv}, we first investigate how information flow is aggregated through attention mechanisms across different layers in four types of tasks, including single- and multi-document QA, summarization, synthetic tasks and code completion. We find that the attention distribution varies for different types of tasks. For example, in summarization tasks, the upper layers require a small KV cache size, while code completion tasks need larger KV cache sizes in the upper layers. This implies that for code completion tasks, upper layers require maintaining a larger KV cache size, in contrast to PyramidKV~\citep{zhang2024pyramidkv}, where the KV cache size decreases as the layer depth increases.

Building on this insight, we propose a task-aware adaptive KV cache compression method, named DynamicKV. Specifically, we first calculate an attention score for the most recent few tokens and all other tokens, which in RAG \citep{lewis2020retrieval} can be viewed as calculating the relevance of the most recent query to the retrieved text. Then, we preset a temporary storage to hold the temporary KV cache states and gradually calculate the size of the final retained temporary storage at each k layer by calculating the size of the correlation mean. It should be noted that at each update, the value is gradually normalized, and the retained temporary storage at each layer is always smaller than the previous one. This temporary storage is determined by the number of tokens that need to be retained, and its size is much smaller than the original cache, thus imposing minimal memory overhead. 
Experiments demonstrate that our DynamicKV can retain full performance while utilizing only 6.9\% of the tokens, and in extreme scenarios, it preserves 90\% of the performance with just 1.7\% of the tokens. Furthermore, experiments on the Needle in a Haystack benchmark revealed that DynamicKV significantly outperforms state-of-the-art (SOTA) methods.

\paragraph{Contributions.} Our main contributions are:

\begin{itemize}
    \item We explore the impact of different task types on token retention at each layer of the LLM. Our findings highlight that for different tasks, token retention varies at each layer, and therefore, dynamic selection of token retention at each layer is necessary for different tasks.
    \item Given our observation, we propose a novel KV cache compression method -- DynamicKV to dynamically adjusts token retention during prefill phase.
    \item Experimental results on the widely used long-context understanding benchmark, LongBench, demonstrate that our approach maintains full performance while using only 6.9\% of the tokens.
\end{itemize}
\section{Related Work}
\textbf{Potential patterns of attention in LLMs.}
The Transformer architecture \citep{vaswani2017attention} has driven progress in NLP through layered refinement of inputs. BERT \citep{devlin2018bert} reveals a hierarchical structure in intermediate layers via \citet{jawahar2019does}: surface features dominate lower layers, evolving into syntactic and semantic representations toward the top. This underscores the capability of LLMs to encode both lexical and complex linguistic information across layers.

For decoder-only models, \citet{fan2024not} demonstrate that intermediate layers suffice for simple tasks, challenging the necessity of full-depth inference. Training strategies like \citep{elhoushi2024layer} further optimize efficiency by introducing layer-wise dropout, enabling early computation exit. Concurrently, KV cache optimization has emerged as a critical direction. \citet{brandon2024reducing} propose Cross-Layer Attention (CLA) to halve cache size via cross-layer attention sharing, while \citet{feng2024adakvoptimizingkvcache} (Ada-KV) dynamically optimize eviction policies by analyzing cross-layer attention patterns. These works highlight the interplay between attention dynamics \citep{feng2024adakvoptimizingkvcache} and memory-efficient computation.
\begin{figure*}[thbp]
    \centering
    \includegraphics[width=1.\linewidth]{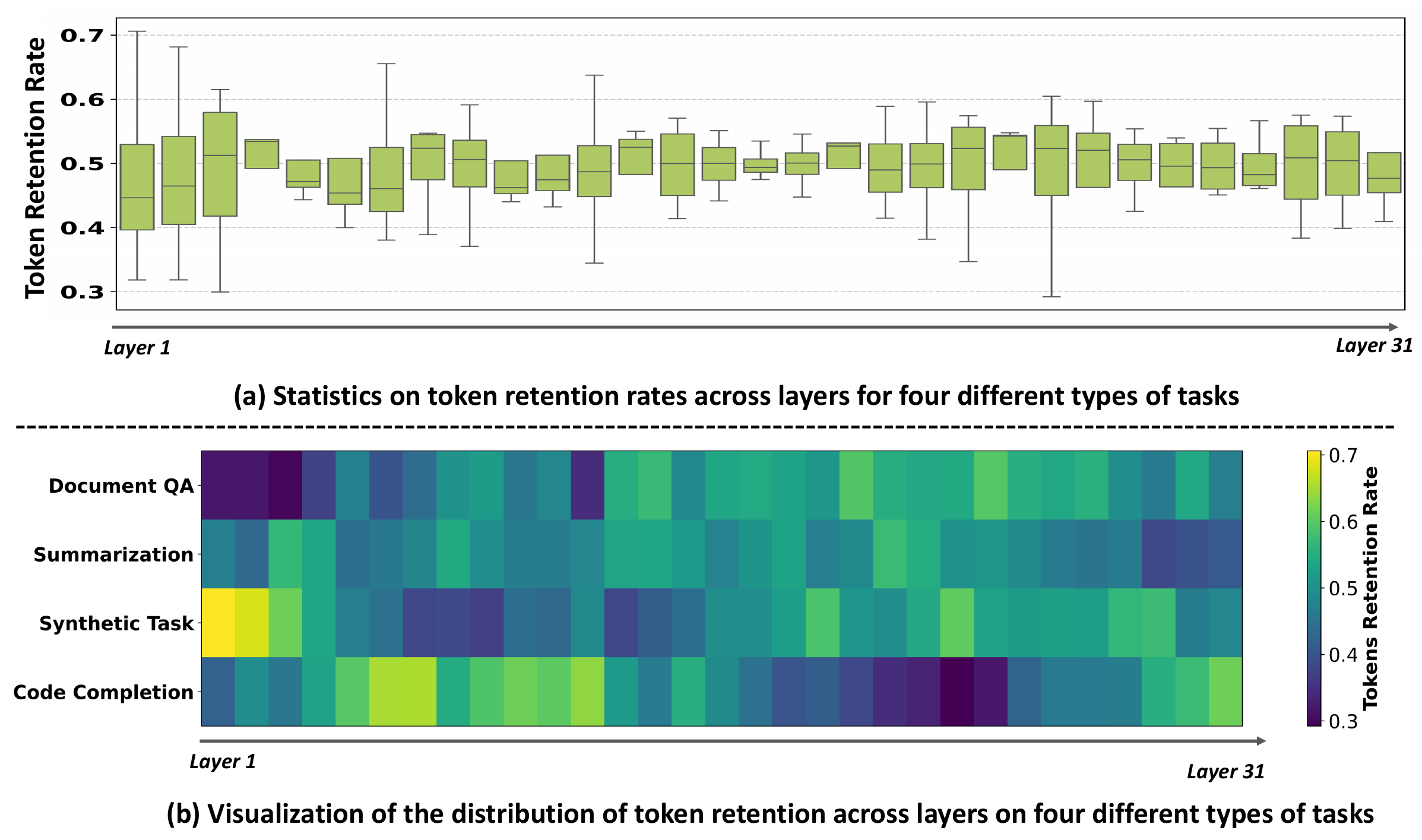}
    \caption{\textbf{Analyzing the distribution of token retention across layers in LlaMA for different tasks, including \textit{Document QA, Summarization, Synthetic Task and Code Completion.}} (a) Each boxplot shows the distribution of token retention rates on different types of tasks across different layers. Results for different layers show that the token retention rates vary significantly across different tasks. (b) We visualize the token retention rates across different layers for four tasks, showing that the token retention rates exhibit different patterns across tasks.}
    \label{fig:observation}
\end{figure*}


\paragraph{Token drop strategies in KV cache compression.} 
Token drop strategies for KV cache compression vary in approach but share a focus on identifying influential tokens. Attention-based methods like FastGen \citep{ge2023model} and Scissorhands \citep{liu2024scissorhands} use attention patterns for pruning. Memory-aware approaches include StreamingLLM \citep{xiao2023efficient}, which prioritizes streaming via attention sinks, and H2O \citep{zhang2024h2o}, which employs cumulative attention scoring for greedy eviction. Hierarchical methods like PyramidKV \citep{zhang2024pyramidkv} adapt by layer but lack generalizability. SnapKV \citep{li2024snapkv} offers task-agnostic compression by selecting key positions per head. Dynamic frameworks such as LazyLLM \citep{fu2024lazyllm} enable flexible token revival, and Ada-KV \citep{feng2024adakvoptimizingkvcache} improves overall performance by optimizing eviction loss bounds over uniform strategies.

Existing methods use fixed patterns across tasks, yet LLMs engage varying layers depending on the task~\cite{elhoushi2024layer}. This suggests token retention during KV cache compression may also differ by task -- an area largely unexplored. This paper examines how task type influences KV cache compression.

\section{Preliminary Studies}
\label{sec:3}
To systematically investigate the attention mechanism across layers in LLMs for long-context inputs, we conduct a fine-grained analysis on four different types of tasks: single- and multi-document question answering (QA), summarization, synthetic tasks, and code completion. 

\paragraph{Experimental setting.}
In particular, we focus our analysis on LlaMA~\citep{dubey2024llama}, visualizing the distribution and behavior of attention across layers to gain deeper insights into its internal mechanisms. Inspired by~\citet{zhang2024pyramidkv}, we calculate the average attention scores between the most recent tokens and all other tokens. Based on these scores, we then identify the top-k (128 multiplied by the number of layers) tokens with the highest attention across all layers. 

\paragraph{Observations.}
As shown in Figure~\ref{fig:observation} (a), we use boxplot to visually present the distribution of four different types of tasks across different layers. We find that different tasks show significantly different token retention rates at a fixed layer. For example, at early layers, the spread is wide, indicating large task-specific variation. To further understand the distribution of token retention rates across different tasks, we visualize the token retention rates across all layers for each task, as shown in Figure 1 (b). We find that \ding{182} \textit{Synthetic Task shows higher retention rates in earlier layers}, \ding{183} \textit{Code Completion shows higher retention rates in the earlier layers as well as the last three layers}, and \ding{184} \textit{Document QA and Summarization exhibit different retention dynamics compared to others}.

\paragraph{Insight.}
The tokens to retain at each layer should adapt dynamically based on the task type.

\section{DynamicKV}
Previous work on KV cache compression~\cite{zhang2024pyramidkv, li2024snapkv} often allocaates a fixed KV cache size across LLM layers. However, as our analysis in \S~\ref{sec:3} demonstrates, attention patterns are not identical across different layers with different types of tasks. Therefore, using fixed KV cache size across layers on different tasks may lead to suboptimal performance. Thus, we propose \textbf{\textit{DynamicKV}}--- a dynamic layer-adaptive KV cache compression method. DynamicKV consists of two steps: (1) Dynamic Budget Allocation and (2) Progressive Cache Update.

\subsection{Dynamic Budget Allocation} 
Traditional token drop methods often prioritize the most recent tokens, as these typically carry the most relevant context for generating the next output. We refer to this set of tokens as the current window, denoted by a window size $ws$. Tokens within this window are given the highest priority for retention. To manage memory efficiently, we first define a maximum KV cache retention budge per layer, denoted $B^l$, calculated as $B^l = (wt-ws) \times r_{max}$, where $r_{max}$ is a scaling ratio and $wt$ is the total number of tokens considered. 

Following the approach of \citet{li2024snapkv}, we guide the selection of remaining tokens (outside the current window) based on their attention scores with respect to the instruction tokens. Tokens with higher attention scores are considered more relevant and are thus prioritized for retention in the GPU cache. 

In a standard LLM, attention is computed as:
\begin{equation}
    A = softmax(\frac{Q\cdot K^T}{\sqrt{d_k}}),
\end{equation}
where $ Q \in \mathbb{R}^{M\times d_{k}} $ and $ K \in \mathbb{R}^{M\times d_{k}} $ are the query and key matrics, respectively, $ d_k $ is the dimensionality of the key/queries, and $ M $ is the sequence length. Inspired by~\citet{li2024snapkv, zhang2024pyramidkv}, we compute per-layer attention scores $A^l$ over the current window using a multi-head pooling operation:
\begin{equation}
    A^l = Pooling(A[:, ws]).
\end{equation}

We then select the top $B^l$ tokens based on the highest values in $A^l$. The corresponding KV states at these positions are retained to form a compressed cache:
\begin{equation}
    KV_{retained}^l = KV^l[arg\ topK(A^l, B^l)].
\end{equation}

\subsection{Progressive Cache Update}
To further reduce KV cache usage in the middle layers, we partition the model into blocks of $m$ consecutive layers. For each such block, we dynamically determine the minimal initial retention threshold required to meet cumulative retention demands, while also refreshing the historical KV cache. At the end of each $m$-layer block, we normalize the retention scores to prioritize operationally critical tokens. This process yields a layer-specific budget allocation $Z'$, which facilitates an efficient and adaptive distribution of the cache budget across layers.
Specifically, we apply a top-K selection to retain the most relevant tokens across these layers, and the compute the retention count per layer using a counting function $\Phi$:
\begin{equation}
C^l = Norm(\frac{1}{n}\cdot \Phi (TopK(A, (wt-ws)\times n),
\end{equation}
where $n$ is the number of progressive update layers processed so far, and $(wt-ws)$ denotes the number of tokens outside the current window.

Next, we compute a provisional budget $Z$ by scaling each layer's retention score relative to the maximum:
\begin{equation}
    Z = \left[ \frac{B^l \times t}{\textit{max}(C^l)} | \ t\in C^l\right],
\end{equation}
where $B^l$ is the per-layer retention budget. This is then normalized across layers to ensure the total budget $B=(wt-ws)\times L$ is respected:
\begin{equation}
    Z^\prime = [k\cdot \frac{B}{\sum Z} | k\in Z].
\end{equation}

\begin{figure*}[!t]
    \centering
\includegraphics[width=1.\textwidth]{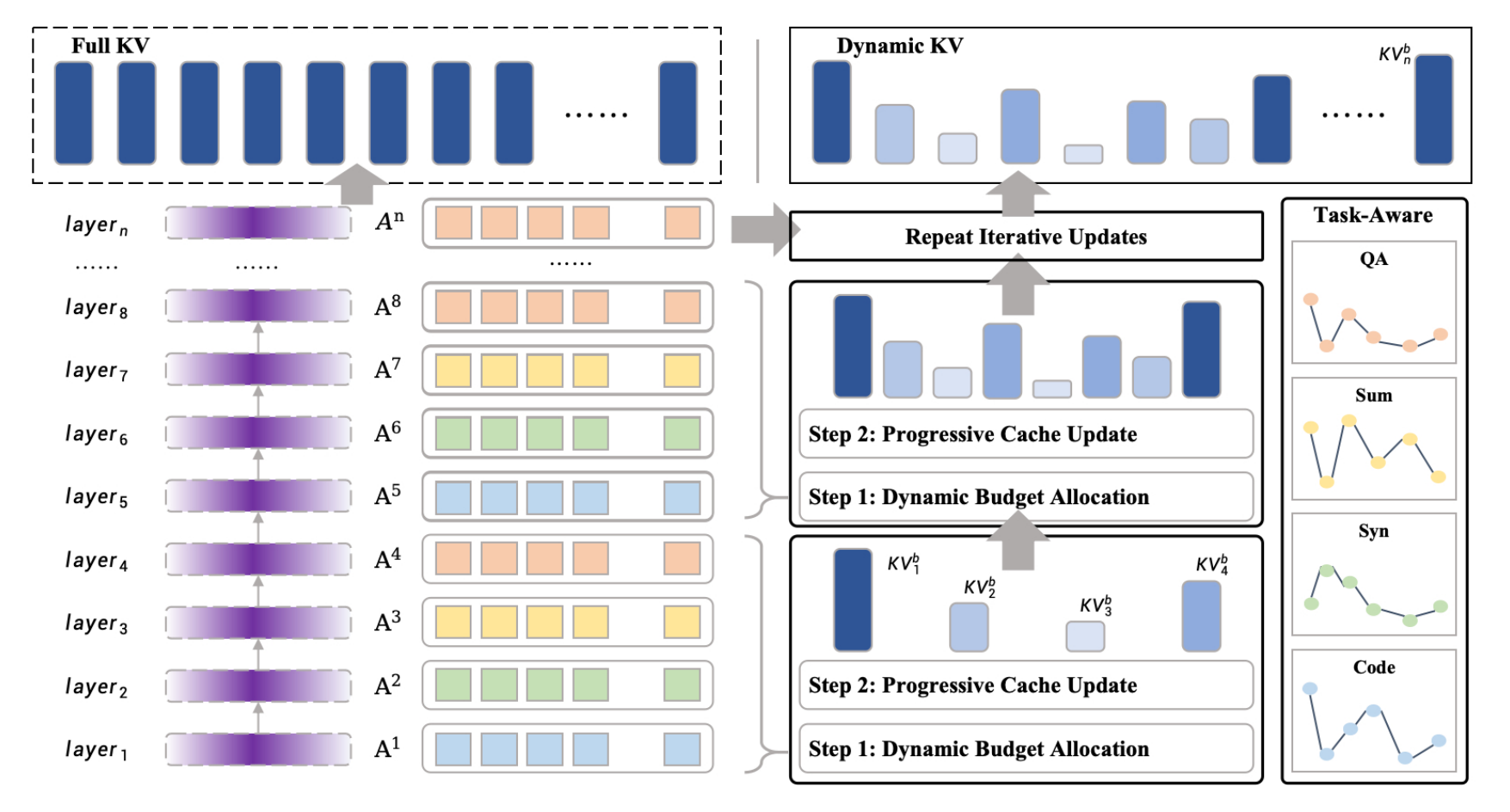}
    \caption{
        \textbf{Overview of our DynamicKV structure and KV cache compression comparison.}
        Left: Layer-wise KV cache retention mechanism in transformer architectures. 
        Right: Our proposed DynamicKV framework employs stage-wise dynamic updating to maintain KV cache within predefined memory budgets, with task-specific visualization showing KV cache preservation patterns across layers.
    }
    \label{figs:model_overview}
\end{figure*}

In practice, during the progressive update of the first $m$ layers, the mechanism uses the attention scores $A$ to estimate the optimal number of tokens to retain per layer. The function $\Phi$ counts the top-K attention entries assigned to each layer, forming $C^l$, which is then normalized into $Z$. Finally, the budget $Z'$ governs how the KV cache is refined for each layer, enabling an adaptive and effective compression strategy across the different layers.

The above process can be expressed as Algorithm \ref{alg: dynamickv}.
\begin{algorithm*}[tb]
    \caption{DynamicKV in Prefill Phase}
    \label{alg: dynamickv}
    \begin{algorithmic}[1]
        \STATE {\bfseries Input:}
            initial budget K/V cache list ${K}^{b}$, ${V}^{b}$,
            ratio max $r_{max}$, 
            update interval $m$, 
            mean token length $wt$, 
            window size $ws$, 
            sequence length $S$, 
            head dimention $d_k$, 
            input embedding of window size ${X}^{ws} \in \mathbb{R}^{ws*d}$,
            initial budget Attention list computed by window token and others ${A}^{b}$,
        \STATE {\bfseries Output:} 
            Compressed K/V cache ${K}^{c}$, ${V}^{c}$
        \STATE $B^l = (wt - ws) \times r_{max}$
        \STATE \textbf{def} \texttt{Update\_Buffer\_Length}($ A $, $ l $):
            \STATE \hskip1.5em ${A}^{gather}$ $\gets$ cat(([${A}$ for $l$ in (1, $l$)]), 0).view(-1)
            \STATE \hskip1.5em $cnts$ $\gets$ Count\_Elemnets(topk(${A}^{gather}$, k=$(wt-ws)*H*l$).indices / ($L*S$)) / $l$
            \STATE \hskip1.5em Compute the $norm$ of $cnts$, range in (0, 1)
            \STATE \hskip1.5em $Z$ $\gets$ [int(($B^l * t$ / max($norm$))) for $t$ in $norm$]
            \STATE \hskip1.5em $r$ $\gets$ sum($Z$) / (($wt-ws$)$*L$)
            \STATE \hskip1.5em $Z^\prime$ $\gets$ [int($k / r$) for $k$ in $Z$]
            \STATE \hskip1.5em Return $Z^\prime$
        
        \FOR{$l \gets 1 \textbf{ to } L$}
            \STATE Compute full KV states ${K}^{s}$, ${V}^{s}$
            \FOR{$h \gets 1 \textbf{ to } H$}
                \STATE \textcolor{gray}{\textbf{\textit{/* compute the Attention between window size token and other all token */}}}
                \STATE ${A}^l_h$ $\gets$ softmax((${X}^{ws}{W}_{h}^{Q}$) $\cdot$ ${K}_{h}^{T}$).mean(dim=-2).pooling(dim=-1)
            \ENDFOR
            \STATE Append ${A}^{l}$ to ${A}^{b}$    \textcolor{gray}{\textbf{\textit{/* current ${A}_{l}$ shape is [$H$, $S$] */}}}
            \STATE \textcolor{gray}{\textbf{\textit{/* calculate current layer buffer KV cache */}}}
            \STATE indices $\gets$ ${A}^{l}$.topk($B^l$, dim=-1).indices.unsqueeze(-1).expand(-1, -1, $d_k$)
            \STATE ${K}_{l}^{b}$ $\gets$ cat((${K}^{s}$[:,:$-ws$,:].gather(dim=-2, indices),${K}^{s}$[:,$-ws$:,:]), dim=-2)
            \STATE ${V}_{l}^{b}$ $\gets$ cat((${V}^{s}$[:,:$-ws$,:].gather(dim=-2, indices),${V}^{s}$[:,$-ws$:,:]), dim=-2)

            \STATE \textcolor{gray}{\textbf{\textit{/* gradually compress*/}}}
            \IF{$l$ $\%$ $m$ == 0}  
                \STATE $Z^\prime$ $\gets$ \texttt{Update\_Buffer\_Length}(${A}^{l}$, $l$)
                \STATE \textcolor{gray}{\textbf{\textit{/* update the buffer K/V Cache*/}}}
                \FOR{$i$ $\gets$ 1 \textbf{to} $l$}
                    \STATE ${K}_{i}^{b}$ $\gets$ cat((${K}_{l}^{b}$[:,:${Z^\prime}_{i}$,:], ${K}_{l}^{b}$[:,$-ws$:,:]), dim=-2)
                    \STATE ${V}_{i}^{b}$ $\gets$ cat((${V}_{l}^{b}$[:,:${Z^\prime}_{i}$,:], ${V}_{l}^{b}$[:,$-ws$:,:]), dim=-2)
                \ENDFOR
            \ENDIF
        \ENDFOR
        \STATE \text{Update the K/V Cache } ${K}^{c},{V}^{c}$ from ${K}^{b},{V}^{b}$
    \end{algorithmic}
\end{algorithm*}

\subsection{Implementation Details}
Durint the inference, the process is divided into two phases, the prefilling phase and the decoding phase, consistent with existing inference engines~\cite{kwon2023efficient}. Our DynamicKV, while potentially encountering sample-specific attention patterns when determining the optimal KV cache size per layer, performs this step during the prefilling phase. During the decoding phase, no modifications are applied.

\paragraph{\textit{Q1: Does the DynamicKV handles batched inference?}}
\itbf{A1: Yes.}
In fact, modern LLM inference and serving engines (\textit{e.g.,} vLLM~\citet{kwon2023efficient}) generally process samples individually (\textit{i.e.,} batch size=$1$) in prefilling phase, while decoding allows for efficient parallel computation in batches. Since our DynamicKV introduces no modifications during decoding, our method aligns seamlessly with existing inference engines, ensuring that the decoding phase remains fully compatible with batched execution for high-throughput generation.

\paragraph{\textit{Q2: How does the DynamicKV compatible with FlashAttention?}}
\itbf{A2: Our DynamicKV can compatible with FlashAttention during the decoding phase.}
Although our DynamicKV modifies the computation of attention scores during the prefilling phase, which limits compatibility with FlashAttention, it remains highly efficient. This is because attention is computed only within a small widow size $ws$, where $ws \ll M$, keeping the overhead minimal even without FlashAttention. In contrast, no modifications are applied in decoding phase, where we take advantage of FlashAttention to significantly improve computational efficiency.

\section{Experiments}
We conduct comprehensive comparative and ablation experiments to verify the effectiveness of our DynamicKV. In 
\S~\ref{sec:5_1}, we introduce the models, datasets and baselines used in our experiments. 
\S~\ref{sec:5_2} provides a performance comparison between DynamicKV and baseline approaches. Next, in \S~\ref{sec:5_5}, we conduct an ablation study on the parameters of our method to validate its feasibility. We presnet the computational overhead in \S~\ref{sec:5_4}. Finally, in \S~\ref{sec:5_3}, we present the results of DynamicKV on the Needle in Haystack Task.

\subsection{Experimental Settings}
\label{sec:5_1}
\textbf{Models and Context Length.} We utilize the official checkpoints of recently released models from huggingface including LlaMA-3-8B-Instruct~\citep{dubey2024llama}, Qwen-2-7B-Instruct~\citep{yang2024qwen2}, Mistral-7B-Instruct-v0.2~\citep{jiang2023mistral}, and InternLM-2.5-7B-Chat-1M~\citep{cai2024internlm2} as our base models, which support context lengths of 8k, 32k, 32k, and 1M tokens respectively.

\paragraph{Datasets.} 
LongBench is a comprehensive benchmark for evaluating the contextual understanding capabilities of LLMs. For our comparative experiments, we use 16 English datasets from this benchmark, specifically NarrativeQA~\citep{kovcisky2018narrativeqa}, Qasper~\citep{dasigi2021dataset}, MultiFieldQA-en, HotpotQA~\citep{yang2018hotpotqa}, 2WikiMultihopQA~\citep{ho2020constructing}, MuSiQue~\citep{trivedi2022musique}, GovReport~\citep{huang2021efficient}, QMSum~\citep{zhong2021qmsum}, MultiNews~\citep{fabbri2019multi}, TREC~\citep{li2002learning}, TriviaQA~\citep{joshi2017triviaqa}, SAMSum~\citep{gliwa2019samsum}, PassageCount, PassageRetrieval-en, LCC~\citep{guo2023longcoder}, and RepoBench-P~\citep{liu2023repobench}.

\paragraph{Baselines.} We evaluate the recent fixed-pattern token-dropping methods, including:
(1) \textbf{StreamingLLM}~\cite{xiao2023efficient}, which utilizes attention sinks and rolling KV caches to retain the most recent tokens.
(2) \textbf{H2O}~\cite{zhang2024h2o}, which employs a Heavy Hitter Oracle for KV cache eviction.
(3) \textbf{SnapKV}~\cite{li2024snapkv}, which selects important tokens for each attention head through clustering.
(4) \textbf{PyramidKV}~\cite{zhang2024pyramidkv}, which introduces a pyramid pattern where layers select important tokens in a monotonically decreasing manner.

\subsection{Comparative Experiments on LongBench}
\label{sec:5_2}

\begin{table*}[t]
\centering
\resizebox{\linewidth}{!}{
\begin{tabular}{l@{\hspace{0.05ex}}|c@{\hspace{0.05ex}}c@{\hspace{0.05ex}}c@{\hspace{0.05ex}}c@{\hspace{0.05ex}}c@{\hspace{0.05ex}}c@{\hspace{0.05ex}}c@{\hspace{0.05ex}}c@{\hspace{0.05ex}}c@{\hspace{0.05ex}}c@{\hspace{0.05ex}}c@{\hspace{0.05ex}}c@{\hspace{0.05ex}}c@{\hspace{0.05ex}}c@{\hspace{0.05ex}}c@{\hspace{0.6ex}}c@{\hspace{0.6ex}}c@{\hspace{0.6ex}}c}

\toprule

\multirow{6}{*}{\rotatebox{30}{\textbf{Model}}}  & \multirow{6}{*}{\rotatebox{30}{\textbf{Method}}}  & \multicolumn{3}{c}{\itbf{Single-Document QA}} & \multicolumn{3}{c}{\itbf{Multi-Document QA}}& \multicolumn{3}{c}{\itbf{Summarization}}& \multicolumn{3}{c}{\itbf{Few-shot Learning}}& \multicolumn{2}{c}{\itbf{Synthetic}} & \multicolumn{2}{c}{\itbf{Code}} & \multirow{6}{*}{\itbf{Avg.}} \\

\cmidrule(lr){3-5}\cmidrule(lr){6-8}\cmidrule(lr){9-11}\cmidrule(lr){12-14}\cmidrule(lr){15-16}\cmidrule(lr){17-18}

 & & \rotatebox[origin=c]{30}{\textbf{NrtvQA}} & \rotatebox[origin=c]{30}{\textbf{Qasper}} & \rotatebox[origin=c]{30}{\textbf{MF-en}} & \rotatebox[origin=c]{30}{\textbf{HotpotQA}} & \rotatebox[origin=c]{30}{\textbf{2WikiMQA}} & \rotatebox[origin=c]{30}{\textbf{Musique}} & \rotatebox[origin=c]{30}{\textbf{GovReport}} & \rotatebox[origin=c]{30}{\textbf{QMSum}} & \rotatebox[origin=c]{30}{\textbf{MultiNews}} & \rotatebox[origin=c]{30}{\textbf{TREC}} & \rotatebox[origin=c]{30}{\textbf{TriviaQA}} & \rotatebox[origin=c]{30}{\textbf{SAMSum}} & \rotatebox[origin=c]{30}{\textbf{PCount}} & \rotatebox[origin=c]{30}{\textbf{PRe}} & \rotatebox[origin=c]{30}{\textbf{Lcc}} & \rotatebox[origin=c]{30}{\textbf{RB-P}} & \\
\cmidrule(lr){3-18} &  &18409 &3619 &4559 &9151 &4887 &11214 &8734 &10614 &2113 &5177 &8209 &6258 &11141 &9289 &1235 &4206& -- \\

\midrule
\multirow{6}{*}{\rotatebox{90}{\textbf{\makecell{LlaMA-3-8B\\-Instruct}}}}
 &FullKV &25.16 &31.81 &39.59 &43.09 &36.15 &21.77 &28.62 &23.34 &26.33 &75.00 &90.50 &42.36 &5.20 &69.25 &59.04 &53.93 &41.95 \\
\cmidrule(lr){2-19} &StreamingLLM &19.03 &12.78 &28.67 &37.83 &29.97 &16.55 &20.30 &20.94 &\textbf{24.56} &61.00 &75.43 &30.82 &5.86 &69.50 &51.93 &49.98 &34.70 \\
&H2O &22.84 &16.80 &32.36 &41.43 &34.07 &19.30 &22.28 &22.81 &23.69 &41.00 &\textbf{90.46} &40.19 &5.54 &69.50 &57.52 &55.43 &37.20 \\
&SnapKV &24.62 &22.78 &\textbf{37.88} &42.96 &\textbf{34.82} &20.65 &22.63 &22.54 &23.93 &70.00 &90.39 &40.30 &5.74 &69.50 &60.27 &55.85 &40.30 \\
&PyramidKV &24.48 &23.51 &36.14 &42.33 &31.95 &20.73 &\textbf{23.37} &\textbf{23.01} &24.37 &72.50 &90.43 &40.54 &\textbf{5.88} &69.50 &59.25 &54.87 &40.18 \\
 & \textbf{Ours} & \textbf{24.78} & \textbf{24.76} &36.84 &\textbf{44.13} &33.25 &\textbf{20.82} &23.00 &22.76 &24.14 &\textbf{72.50} &90.39 &\textbf{40.76} &5.78 &\textbf{69.50} & \textbf{61.40} &\textbf{56.91} &\textbf{40.73} \\

\midrule

\multirow{6}{*}{\rotatebox{90}{\textbf{\makecell{Mistral-7B\\-Instruct-v0.2}}}}
 &FullKV &26.63 &32.99 &49.34 &42.77 &27.35 &18.77 &32.87 &24.24 &27.10 &71.00 &86.23 &42.96 &2.75 &86.98 &56.93 &54.49 &42.71 \\
\cmidrule(lr){2-19} &StreamingLLM &19.05 &17.21 &36.82 &30.64 &21.84 &10.56 &24.47 &19.84 &\textbf{25.48} &62.00 &72.82 &29.49 &2.71 &19.25 &46.15 &42.55 &30.06 \\
&H2O &22.33 &25.75 &44.09 &32.76 &22.88 &14.96 &23.53 &22.96 &24.53 &41.50 &85.53 &41.54 &3.39 &86.20 &55.11 &50.81 &37.37 \\
&SnapKV &24.95 &27.97 &\textbf{49.04} &39.93 &25.18 &\textbf{17.64} &24.14 &23.69 &24.47 &67.50 &86.04 &41.14 &2.90 &\textbf{86.98} &\textbf{56.73} &\textbf{53.11} &40.71 \\
&PyramidKV &23.49 &28.79 &48.71 &\textbf{41.00} &25.64 &16.35 &\textbf{24.79} &23.52 &24.49 &69.50 &86.20 &42.58 &\textbf{3.53} &81.81 &55.45 &51.67 &40.47 \\
& \textbf{Ours} &\textbf{25.63} &\textbf{29.11} &48.41 &39.85 &\textbf{26.62} &16.72 &24.73 &\textbf{23.72} &24.83 &\textbf{70.50} &\textbf{86.74} &\textbf{43.01} &3.20 &83.57 &55.40 &52.35 &\textbf{40.90} \\

\midrule

\multirow{6}{*}{\rotatebox{90}{\textbf{\makecell{Qwen2-7B\\-Instruct}}}}
 &FullKV &25.14 &42.35 &45.04 &14.80 &14.13 &9.23 &36.35 &23.79 &26.51 &76.50 &89.16 &45.23 &6.50 &75.50 &60.30 &60.78 &40.71 \\
\cmidrule(lr){2-19} &StreamingLLM &20.47 &26.97 &32.64 &14.31 &14.39 &6.82 &\textbf{25.70} &19.31 &\textbf{24.88} &66.00 &76.56 &32.11 &\textbf{8.00} &15.50 &46.58 &44.20 &29.65\\
&H2O &22.88 &34.28 &41.40 &13.30 &14.60 &8.31 &23.69 &22.07 &22.72 &39.50 &88.75 &\textbf{43.91} &6.00 &72.00 &58.83 &57.83 &35.63 \\
&SnapKV &23.86 &38.61 &44.65 &\textbf{15.60} &\textbf{14.62} &\textbf{9.13} &24.56 &22.39 &23.07 &70.00 &89.31 &43.32 &5.00 &72.00 &58.67 &\textbf{60.74} &38.47 \\
&PyramidKV &24.47 &37.60 &43.51 &14.48 &12.83 &8.99 &23.59 &22.30 &22.41 &74.00 &89.21 &43.40 &6.50 &74.00 &57.67 &56.14 &38.19 \\
& \textbf{Ours} &\textbf{24.66} &\textbf{40.44} &\textbf{45.30} &15.42 &13.89 &8.46 &25.51 &\textbf{22.77} &22.92 &\textbf{74.00} &\textbf{89.27} &43.18 &7.00 &\textbf{74.00} &\textbf{60.38} &59.33 &\textbf{39.16} \\

\midrule

\multirow{6}{*}{\rotatebox{90}{\textbf{\makecell{InternLM-2.5-7B\\-Chat-1M}}}}
 &FullKV &22.42 &27.61 &39.98 &40.92 &33.48 &26.68 &33.01 &25.18 &26.28 &72.50 &86.76 &39.76 &2.91 &100.00 &55.86 &57.95 &43.21 \\
\cmidrule(lr){2-19} &StreamingLLM &17.58 &15.86 &26.55 &26.68 &16.69 &11.01 &\textbf{25.96} &21.33 &\textbf{25.57} &65.00 &67.16 &21.71 &0.95 &87.56 &43.58 &42.76 &32.25 \\
&H2O &15.33 &19.84 &32.41 &27.88 &20.10 &21.13 &16.91 &22.99 &21.49 &41.00 &84.38 &34.76 &\textbf{1.23} &96.50 &48.46 &50.00 &34.65 \\
&SnapKV &16.86 &23.28 &36.24 &32.14 &19.89 &\textbf{23.21} &17.69 &23.18 &22.44 &71.00 &84.05 &34.34 &1.00 &\textbf{96.50} &50.32 &\textbf{53.34} &37.84 \\
&PyramidKV &17.62 &21.08 &37.52 &32.21 &\textbf{21.31} &22.03 &19.37 &\textbf{24.06} &22.22 &73.00 &83.94 &34.61 &1.05 &95.50 &50.45 &49.72 &37.86 \\
&  \textbf{Ours} &\textbf{17.77} &\textbf{23.87} &\textbf{37.74} &\textbf{32.98} &21.13 &20.85 &19.13 &23.49 &22.48 &\textbf{75.00} &\textbf{84.89} &\textbf{36.70} &0.91 &95.50 &\textbf{50.70} &51.08 &\textbf{38.39} \\

\bottomrule
\end{tabular}
}
\caption{\textbf{Performance comparison on the LongBench dataset} for full KV cache, previous methods (StreamingLLM, H2O, SnapKV, PyramidKV), and our DynamicKV method, with KV cache sizes of 512, using models including LLaMA3-8B-Instruct, Mistral-7B-Instruct-v0.2, QWen2-7B-Instruct, and InternLM-2.5-Chat-1M. Bold indicates the best performance.}
\label{table:longbench}
\end{table*}

With the total KV cache size constrained to just 512, we evaluate the performance retention of StreamingLLM, H2O, SnapKV, PyramidKV, and our proposed approach, DynamicKV, relative to the FullKV. As shown in Table~\ref{table:longbench}, DynamicKV consistently outperforms existing methods, enven when operating with an exceptionally low cache-to-context ratio of only 6.9\%. Notably, DynamicKV exceeds the best-performing baseline by 0.43\%, 0.19\%, 0.69\%, and 0.53\% across comparable models -- retaining 97\%, 96\%, 96\%, and 89\% of FullKV's performance, respectively. These results underscore DynamicKV's remarkable ability to preserve near FullKV-level performance under extreme memory constraints. Further more, DynamicKV not only matches but enhances PyramidKV's capabilities on complex tasks such as code completion, significantly extending the performance ceiling at lower cache capacities. In addition, we also compared the performance with a KV cache size of 128. The detailed results can be found in Appendix~\ref{appendix_longbench_128}.

\subsection{Ablation Study}
\label{sec:5_5}
\begin{figure}[hpt]
		\begin{center}
			\includegraphics[width=1.\linewidth]{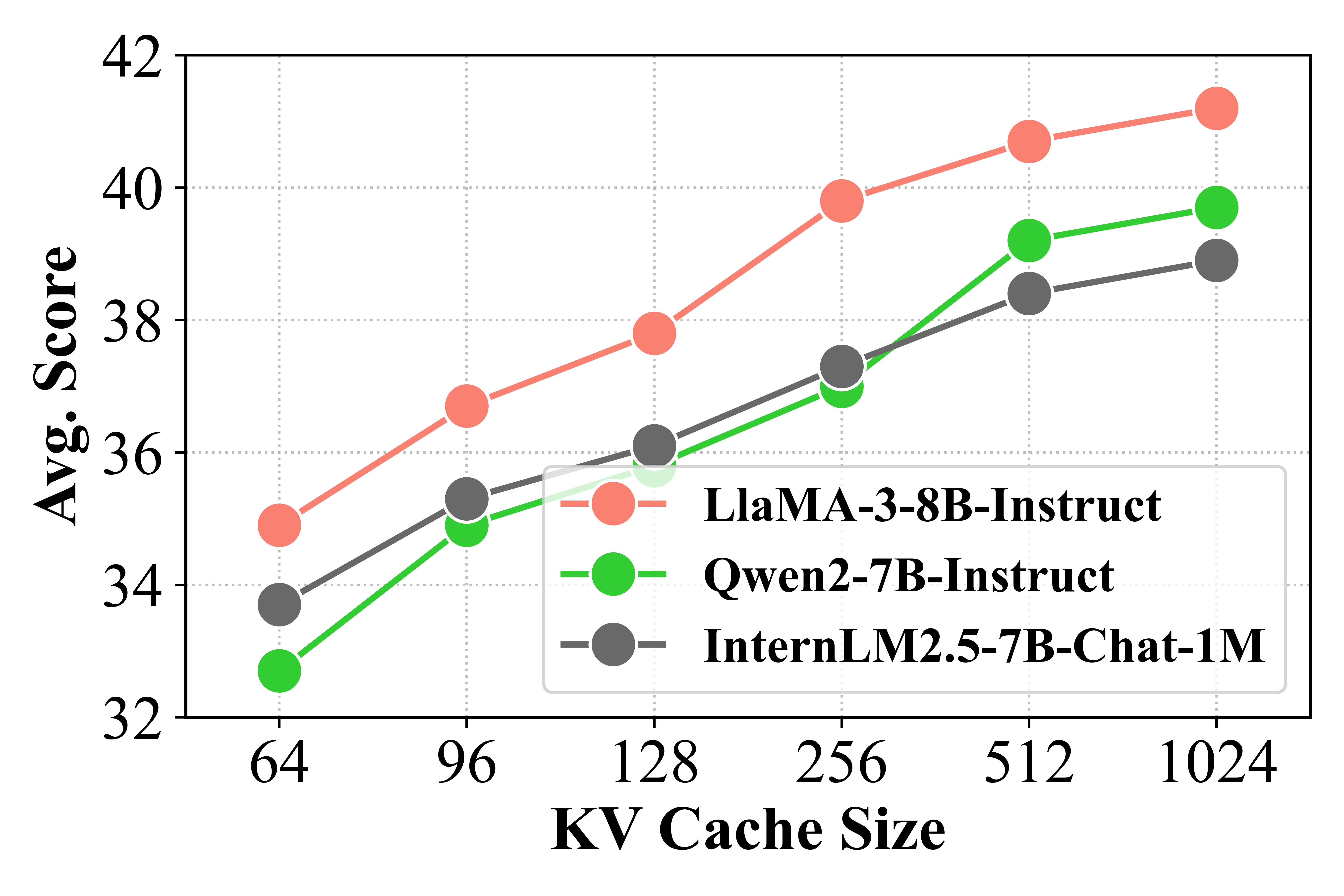}
		\end{center}
		\caption{\textbf{Performance of DynamicKV with different KV cache size on LongBench.} The evaluation metrics are the average score of LongBench across datasets. }
		\label{fig:kv_cache}
		\vspace{-0.4cm}
	\end{figure}

In this study, we investigate the performance of the DynamicKV mechanism across varying key-value cache sizes. 
The results, as shown in Figure~\ref{fig:kv_cache}, reveal a consistent improvement in performance with an increase in the cache size for all evaluated models. For the LlaMA-3-8B-Instruct, the performance metric improved from 34.93 to 41.22 as the key-value cache size was increased from 64 to 1024. This improvement is also applicable to other models.
These findings underscore the effectiveness of the DynamicKV cache in leveraging KV cache compression to maintain the capabilities of long context. Notably, a larger cache capacity is generally associated with superior performance. Nonetheless, it is essential to strike a balance when selecting the cache size, taking into account the practical constraints related to storage and computational resources.

\subsection{Computational Overhead}
\label{sec:5_4}
To better understand the overhead of our DynamicKV, we compare the computational overhead with the FullKV using Llama on LongBench. The evaluation metrics are \textbf{Time-to-First-Token (TTFT), Time-Per-Output-Token (TPOT), end-to-end latency}, and \textbf{GPU memory usage (GB)}.
We present the result in Table~\ref{table:efficiency}.

We can observe that DynamicKV deliver 129\% higher TPOT, 56\% lower latency comparison with FullKV. Experimental results show that \itbf{our DynamicKV offers significant advantages in both computational efficiency and memory usage.} More efficient experimental results can be found in Appendix~\ref{appendix_efficiency}.

\begin{table}[htp]
\centering
\resizebox{1.\linewidth}{!}{
\begin{tabular}{lccccc}
\toprule
\textbf{Method}    & \textbf{TTFT}$\uparrow$    & \textbf{TPOT}$\uparrow$ & \textbf{Latency}$\downarrow$ & \textbf{Memory}$\downarrow$ \\ \hline
FullKV    &  3.52   & 11.65 & 706.56 & 30.48  \\
 \textbf{DynamicKV} & \textbf{3.58} & \textbf{26.69} & \textbf{310.56} & \textbf{27.06} \\ \bottomrule
\end{tabular}
}
\caption{\textbf{Efficiency comparison between FullKV and DynamicKV.} We conduct experiments with a fixed context window ($m=128$), the input length is $32$K and output length is $8$K.}
\label{table:efficiency}
\end{table}

\subsection{Visualization on Needle-in-Haystack Task}
\label{sec:5_3}
\begin{figure*}[t]
    \centering
    \begin{subfigure}[b]{1.\textwidth}
        \includegraphics[width=\textwidth]{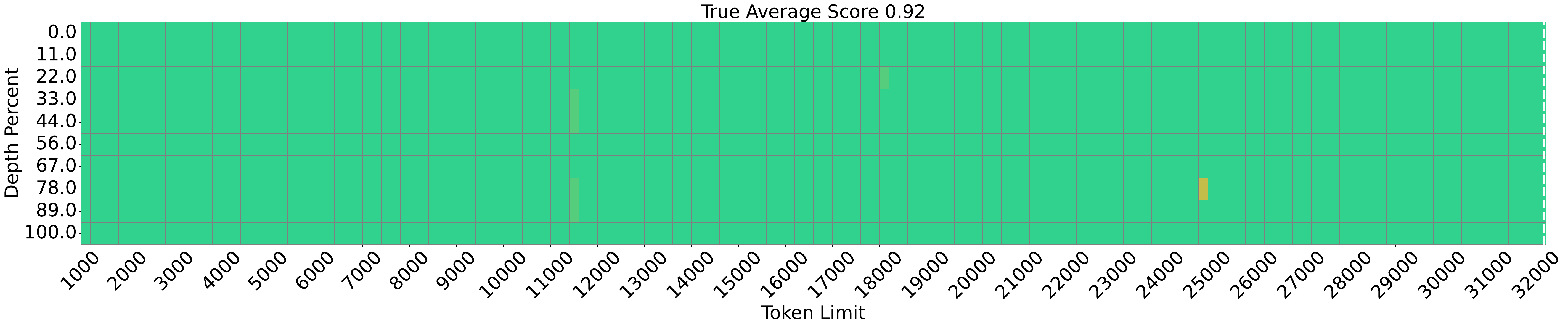}
        \caption{FullKV}
        \label{fig:sub1}
    \end{subfigure}
    \begin{subfigure}[b]{1.\textwidth}
        \includegraphics[width=\textwidth]{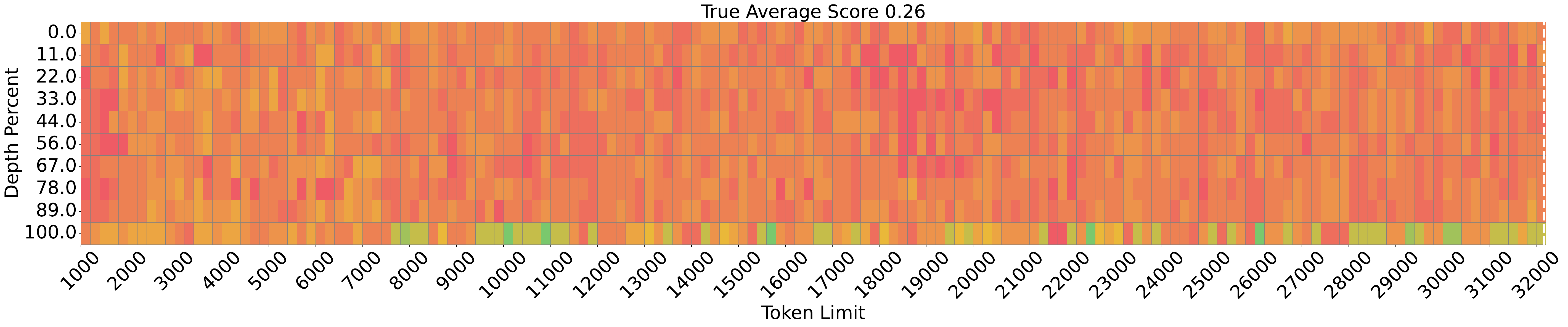}
        \caption{StreamingLLM}
        \label{fig:sub2}
    \end{subfigure}
    \begin{subfigure}[b]{1.\textwidth}
        \includegraphics[width=\textwidth]{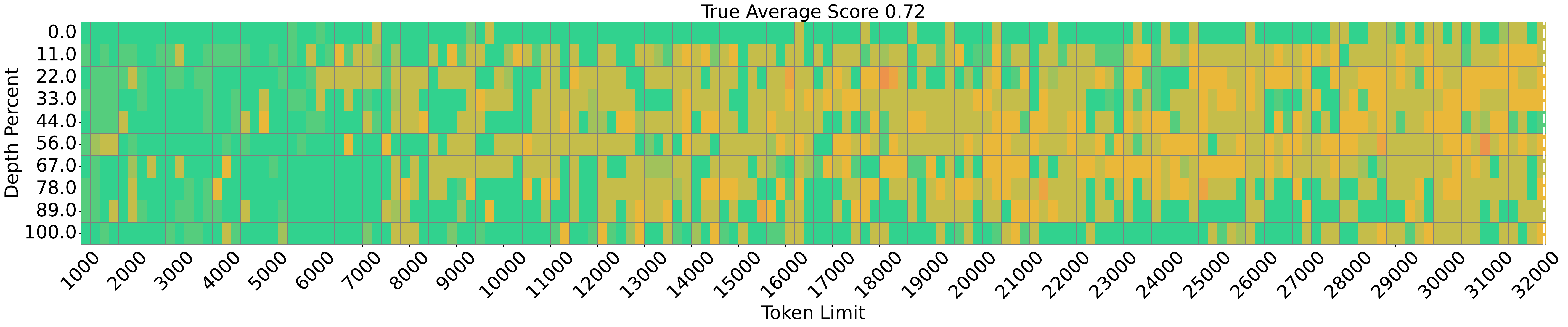}
        \caption{PyramidKV}
        \label{fig:sub5}
    \end{subfigure}
    \begin{subfigure}[b]{1.\textwidth}
        \includegraphics[width=\textwidth]{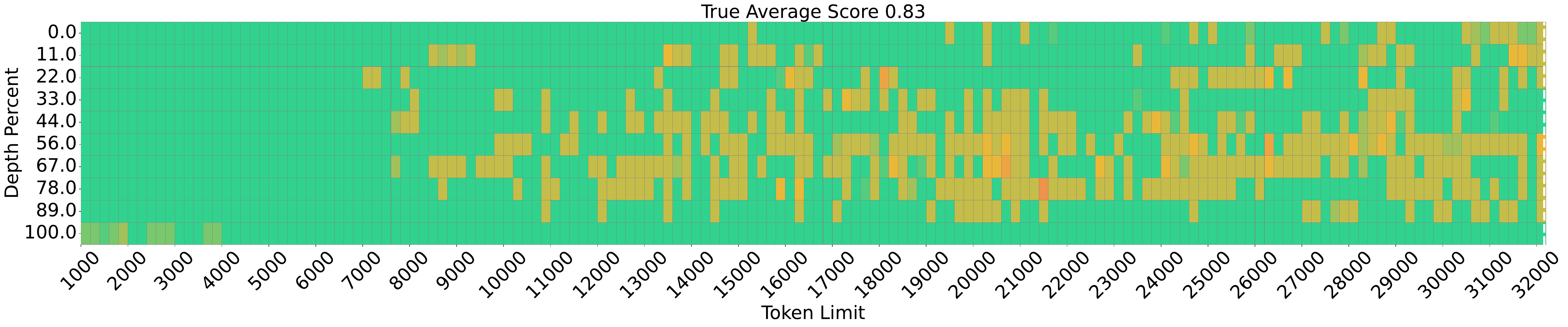}
        \caption{DynamicKV}
        \label{fig:sub6}
    \end{subfigure}
    \caption{\textbf{Performance Comparison on the Needle in a Haystack Task} using Mistral-7B-Instruct-v0.2 with 32k context size in 64 KV cache size. The vertical axis of the table represents the depth percentage, and the horizontal axis represents the length.}
    \label{fig:needle_in_haystack_mistral}
\end{figure*}
We evaluate the in-context retrieval capabilities of LLMs using the ``Fact Retrieval Across Context Lengths'' benchmark (also known as \textit{Needle In A Haystack}) -- a challenging dataset designed to assess whether a model can accurately extract key information from long input sequences. To this end, we adopt Mistral as the base model and extend the context length up to 32K tokens. We compare multiple KV cache compression strategies, including StreamingLLM, PyramidKV, and our proposed DynamicKV, at cache sized of 64 and the FullKV baseline. The results, shown in Figure~\ref{fig:needle_in_haystack_mistral}, highlight that DynamicKV retains 90\% of the model's original performance even under aggressive compression -- achieving accuracy gains of 57\%, 37\%, 41\% and 11\% over competing methods.

Moreover, the results demonstrate that at context lengths up to 7K tokens, DynamicKV's extreme compression nearly achieves full accuracy. Beyond this range, it continues to significantly outperform all baselines. These results underscore DynamicKV's superior capability in hierarchical token selection, and validate our hypothesis that \itbf{the distribution of critical tokens across layers is inherently dynamic.}

\subsection*{\ding{43} A Note on More Details in the Appendix}
See Appendix~\ref{appendix_model} and \ref{appendix_dataset} for a more detailed description of the experimental settings, Appendix~\ref{appendix_naih} for additional results from Need in a HayStack, Appendix~\ref{appendix_efficiency} for efficiency experiments and Appendix~\ref{appendix_longbench_128} for result of KV cache size of 128 on the LongBench dataset.

\section{Conclusion}
We investigate task-specific attention patterns in LLMs processing long-context inputs and find distinct attention distributions across tasks. To address this, we propose DynamicKV, a layer-adaptive KV cache compression framework that dynamically optimizes KV cache allocation per layer.
We evaluate the effectiveness and generalizability of DynamicKV through experiments on 16 datasets from the LongBench benchmark, demonstrating its broad applicability and performance benefits. From the results, we mainly conclude that: (1) a wave-like pattern is followed in complex reasoning tasks (e.g., \emph{code completion} tasks); (2) a pyramid-like pattern is followed in \emph{Synthetic} and \emph{Summarization} tasks; (3) The dynamic hierarchical adaptive DynamicKV approach is capable of formulating a relatively appropriate KV cache retention strategy in accordance with diverse tasks. Particularly, in the circumstance of maintaining an extremely small KV cache size, the effect is significantly enhanced. In the future, we hope that there is a more suitable method to perform KV cache compression without increasing the computation.

\section*{Limitations}
Our work has several potential limitations.
First, given the limited computational budget, we only validate our DynamicKV on models Scaling up to super-large model sizes (e.g., 70B), and applying DynamicKV to more cutting-edge model architectures will be more convincing model architectures.
Second, although we have conducted experiments on multiple tasks including single- and multi-document QA, summarization, synthetic tasks, and code completion, the generalization ability of DynamicKV to other tasks or datasets has not been fully explored. Future work will focus on expanding the application scope of DynamicKV to more diverse tasks and datasets.

%

\bibliography{anthology}

\appendix
\newpage
\section{Appendix}
This appendix presents a detailed description of the used models and dataset (Appendix~\ref{appendix_model} and \ref{appendix_dataset}), along with additional results from \itbf{Need in a HayStack} (Appendix~\ref{appendix_naih}), \itbf{comprehensive efficiency experiments} (Appendix~\ref{appendix_efficiency}), and \itbf{more experimenet results on LongBench} (Appendix~\ref{appendix_longbench_128}).

\subsection{Model Details}
\label{appendix_model}
Our experiments are based on four representative open-sourced LLMs, namely LlaMA-3-8B-Instruct, Mistral-7B-Instruct-v0.2, Qwen2-7B-Instruct, and InternLM2.5-Chat-1M. Testing examples are evaluated in a generative format, with answers generated by greedy decoding across all tasks to ensure a fair comparison.
All the model structures and details in our experiment are shown in Table \ref{table:appendix_models}.
\begin{table*}[htbp]
    \centering
    \begin{tabular}{l|cccc}
        \toprule
            \textbf{Configuration} & \makecell{\textbf{LlaMA-3-8B-}\\\textbf{Instruct}} & \makecell{\textbf{Mistral-7B-}\\\textbf{Instruct-v0.2}} & \makecell{\textbf{Qwen2-7B-}\\\textbf{Instruct}} & \makecell{\textbf{InternLM2.5-7B-}\\\textbf{Chat-1M}} \\
        \midrule
            Hidden Size & 4,096 & 4,096 & 3,584 & 4096 \\
            \# Layers   & 32 & 32 & 28 & 32 \\
            \# Query Heads & 32 & 32 & 28 & 32 \\
            \# KV Heads & 8 & 8 & 4 & 8 \\
            Head Size & 128 & 128 & 128 & 128 \\
            Intermediate Size & 14,336 & 14,336 & 18,944 & 14336 \\
            Embedding & False & False & False & False \\
            Vocabulary Size & 128,256 & 32,000 & 151,646 & 92,544 \\
        \bottomrule
    \end{tabular}
    \caption{Configuration of Models.}
    \label{table:appendix_models}
\end{table*}

\begin{table*}[htbp]
\centering  
    \resizebox{0.95\textwidth}{!}{
        \begin{tabular}{lclrccc}
            \toprule
            \textbf{Dataset} & \textbf{Source} & \textbf{Avg length} & \textbf{Metric} & \textbf{Language} & \textbf{\#data} \\
            \midrule
            \emph{Single-Document QA} \\
            NarrativeQA & Literature, Film & 18,409 & F1 & English & 200 \\
            Qasper & Science & 3,619 & F1 & English & 200 \\
            MultiFieldQA-en & Multi-field & 4,559 & F1 & English & 150 \\
            \midrule
            \emph{Multi-Document QA} \\
            HotpotQA & Wikipedia & 9,151 & F1 & English & 200 \\
            2WikiMultihopQA & Wikipedia & 4,887 & F1 & English & 200 \\
            MuSiQue & Wikipedia & 11,214 & F1 & English & 200 \\
            \midrule
            \emph{Summarization} \\
            GovReport & Government report & 8,734 & Rouge-L & English & 200 \\
            QMSum & Meeting & 10,614 & Rouge-L & English & 200 \\
            MultiNews & News & 2,113 & Rouge-L & English & 200 \\
            \midrule
            \emph{Few-shot Learning} \\
            TREC & Web question & 5,177 & Accuracy (CLS) & English & 200 \\
            TriviaQA & Wikipedia, Web & 8,209 & F1 & English & 200 \\
            SAMSum & Dialogue & 6,258 & Rouge-L & English & 200 \\
            \midrule
            \emph{Synthetic Task} \\
            PassageCount & Wikipedia & 11,141 & Accuracy (EM) & English & 200 \\
            PassageRetrieval-en & Wikipedia & 9,289 & Accuracy (EM) & English & 200 \\
            \midrule
            \emph{Code Completion} \\
            LCC & Github & 1,235 & Edit Sim & Python/C\#/Java & 500 \\
            RepoBench-P & Github repository & 4,206 & Edit Sim & Python/Java & 500 \\
            \bottomrule
        \end{tabular}
    }
\caption{An overview of the dataset statistics in LongBench.}
\label{table:appendix_datasets}
\vspace{-5mm}
\end{table*}
\subsection{Dataset Details}
\label{appendix_dataset}
We evaluate the performance of DynamicKV on long-context tasks using LongBench~\cite{bai2023longbench}, a rigorously constructed benchmark suite designed to challenge language models with extended documents and intricate information sequences. Developed for comprehensive, multi-task assessment, LongBench serves as a critical tool for measuring a model's ability to understand and reason over long-context inputs with precision and depth.
The data sources, average length, evaluation metrics, language, and data volume of subdatasets of LongBench are shown in Table~\ref{table:appendix_datasets}.
\begin{table*}[htbp]
    \centering
    \begin{tabular}{l|ccccc}
        \toprule
            \textbf{Model} & \textbf{StreamingLLM} & \textbf{H2O} & \textbf{SnapKV} & \textbf{PyramidKV} & \textbf{DynamicKV} \\
        \midrule
            LlaMA-3-8B-Instruct & 0.29 & 0.46 & 0.80 & 0.89 &  0.9\\
            Qwen-2-7B-Instruct & 0.22 & 0.41 & 0.84 & 0.86 &  0.87\\
        \bottomrule
    \end{tabular}
    \caption{Comparison of different KV cache compression methods in the Needle in a Haystack task.}
    \label{table:appendix_niah}
\end{table*}

\subsection{Need in a HayStack}
\label{appendix_naih}
As shown in Table~\ref{table:appendix_niah}, we compare the performance of various KV cache compression methods -- StreamingLLM, H2O, SnapKV, PyramidKV, and DynamicKV -- on the Needle in a Haystack task using two models: LlaMA-3-8B-Instruct and Qwen-2-7B-Instruct. Across both models, our DynamicKV achieves the highest performance, scoring 0.9 for LlaMA-3-8B-Instruct and 0.87 for Qwen-2-7B-Instruct. These results highlight DynamicKV's superior ability to retain task-critical information in long-context scenarios.

\begin{table*}[tbp]
    \centering
    \begin{tabular}{ccccccc}
        \toprule
            \textbf{Input Len} & \textbf{Output Len} & \textbf{Method} & \textbf{TTFT (s)} & \textbf{TPOT (tok/s)} & \textbf{Latency (s)} & \textbf{Memory (MB)} \\
        \midrule
            8k & 2k & FullKV & 0.66 & 27.63 & 74.79 & 20055 \\
            8k & 2k & Dynamickv & 0.70 & 33.85 & 61.21 & 19417 \\
        \cmidrule(r){1-7}
            16k & 4k & FullKV & 1.45 & 19.55 & 209.56 & 23859 \\
            16k & 4k & Dynamickv & 1.49 & 33.02 & 125.52 & 22051 \\
        \cmidrule(r){1-7}
            32k & 8k & FullKV & 3.52 & 11.65 & 706.56 & 31213 \\
            32k & 8k & Dynamickv & 3.58 & 26.69 & 310.56 & 27713 \\
        \bottomrule
    \end{tabular}
    \caption{Efficiency comparison between FullKV and DynamicKV}
    \label{table:appendix_efficiency}
\end{table*}
\subsection{Efficiency Experiments}
\label{appendix_efficiency}
We evaluate the efficiency of DynamicKV against the standard method (FullKV) under varying input/output lengths. All experiments are conducted with a fixed context window ($m=128$), measuring Time-to-First-Token (TTFT), Time-Per-Output-Token (TPOT), end-to-end latency, and GPU memory usage. The results are summarized in Table~\ref{table:appendix_efficiency}.

Key observations include:

\begin{itemize}
\item \textbf{Short Sequences (8k/2k):} DynamicKV improves TPOT by 22.5\% (27.63→33.85 tok/s) while slightly increasing TTFT by 6\% (0.66s→0.70s), achieving 18.2\% lower total latency (74.79s→61.21s) with 638MB memory reduction.

\item \textbf{Long Sequences (32k/8k):} The advantages amplify significantly, with DynamicKV delivering 129\% higher TPOT (11.65→26.69 tok/s), 56\% lower latency (706.56s→310.56s), and 11.2\% memory savings (31213MB→27713MB).

\item \textbf{Scalability:} FullKV shows superlinear TPOT degradation (11.65 tok/s at 32k inputs), while DynamicKV maintains stable throughput through on-demand computation, demonstrating better adaptability to long-context generation.
\end{itemize}

The experiments demonstrate that dynamic KV caching trades marginal initial latency for substantially better sustained generation speed and memory efficiency, particularly beneficial for long-text generation tasks (>2k output tokens).

\subsection{More Experiment Result on LongBench}
\label{appendix_longbench_128}
\begin{table*}[t]
\centering
\resizebox{\textwidth}{!}{
\begin{tabular}{l@{\hspace{0.05ex}}|c@{\hspace{0.05ex}}c@{\hspace{0.05ex}}c@{\hspace{0.05ex}}c@{\hspace{0.05ex}}c@{\hspace{0.05ex}}c@{\hspace{0.05ex}}c@{\hspace{0.05ex}}c@{\hspace{0.05ex}}c@{\hspace{0.05ex}}c@{\hspace{0.05ex}}c@{\hspace{0.05ex}}c@{\hspace{0.05ex}}c@{\hspace{0.05ex}}c@{\hspace{0.05ex}}c@{\hspace{0.6ex}}c@{\hspace{0.6ex}}c@{\hspace{0.6ex}}c}

\toprule

\multirow{6}{*}{\rotatebox{30}{\textbf{Model}}}  & \multirow{6}{*}{\rotatebox{30}{\textbf{Method}}}  & \multicolumn{3}{c}{\itbf{Single-Document QA}} & \multicolumn{3}{c}{\itbf{Multi-Document QA}}& \multicolumn{3}{c}{\itbf{Summarization}}& \multicolumn{3}{c}{\itbf{Few-shot Learning}}& \multicolumn{2}{c}{\itbf{Synthetic}} & \multicolumn{2}{c}{\itbf{Code}} & \multirow{6}{*}{\itbf{Avg.}} \\

\cmidrule(lr){3-5}\cmidrule(lr){6-8}\cmidrule(lr){9-11}\cmidrule(lr){12-14}\cmidrule(lr){15-16}\cmidrule(lr){17-18}

 & & \rotatebox[origin=c]{30}{\textbf{NrtvQA}} & \rotatebox[origin=c]{30}{\textbf{Qasper}} & \rotatebox[origin=c]{30}{\textbf{MF-en}} & \rotatebox[origin=c]{30}{\textbf{HotpotQA}} & \rotatebox[origin=c]{30}{\textbf{2WikiMQA}} & \rotatebox[origin=c]{30}{\textbf{Musique}} & \rotatebox[origin=c]{30}{\textbf{GovReport}} & \rotatebox[origin=c]{30}{\textbf{QMSum}} & \rotatebox[origin=c]{30}{\textbf{MultiNews}} & \rotatebox[origin=c]{30}{\textbf{TREC}} & \rotatebox[origin=c]{30}{\textbf{TriviaQA}} & \rotatebox[origin=c]{30}{\textbf{SAMSum}} & \rotatebox[origin=c]{30}{\textbf{PCount}} & \rotatebox[origin=c]{30}{\textbf{PRe}} & \rotatebox[origin=c]{30}{\textbf{Lcc}} & \rotatebox[origin=c]{30}{\textbf{RB-P}} & \\
\cmidrule(lr){3-18}

 & &18409 &3619 &4559 &9151 &4887 &11214 &8734 &10614 &2113 &5177 &8209 &6258 &11141 &9289 &1235 &4206& -- \\

\midrule
\multirow{6}{*}{\rotatebox{90}{\textbf{\makecell{LlaMA-3-8B\\-Instruct}}}}
 &FullKV &25.16 &31.81 &39.59 &43.09 &36.15 &21.77 &28.62 &23.34 &26.33 &75.00 &90.50 &42.36 &5.20 &69.25 &59.04 &53.93 &41.95 \\
\cmidrule(lr){2-19}
&StreamingLLM &17.85 &9.50 &23.09 &37.84 &29.02 &16.77 &17.91 &20.42 &20.16 &44.00 &73.00 &30.00 &5.80 &69.50 &48.38 &49.31 &32.03 \\
&H2O &21.58 &12.54 &28.49 &37.13 &32.36 &18.88 &20.23 &22.16 &21.14 &39.00 &86.62 &39.19 &5.50 &69.50 &57.39 &54.46 &35.39 \\
&SnapKV &21.71 &12.37 &32.38 &37.44 &30.48 &19.50 &19.06 &21.36 &20.07 &45.5 &87.74 &38.15 &5.50 &68.85 &57.42 &54.61 &35.76 \\
&PyramidKV &22.26 &16.65 &30.73 &38.97 &29.28 &19.19 &19.92 &22.06 &20.87 &68.00 &88.95 &38.23 &5.92 &69.50 &57.20 &51.54 &37.45 \\
&ours &22.10 &14.93 &32.94 &41.06 &27.98 &21.18 &20.03 &22.06 &21.28 &65.50 &89.61 &38.70 &5.13 &69.50 &58.01 &54.00 &\textbf{37.75} \\

\midrule

\multirow{6}{*}{\rotatebox{90}{\textbf{\makecell{Mistral-7B\\-Instruct-v0.2}}}}
 &FullKV &26.63 &32.99 &49.34 &42.77 &27.35 &18.77 &32.87 &24.24 &27.10 &71.00 &86.23 &42.96 &2.75 &86.98 &56.93 &54.49 &42.71 \\
\cmidrule(lr){2-19}
&StreamingLLM &16.58 &14.76 &30.36 &28.13 &21.76 &11.98 &18.26 &19.02 &19.16 &43.50 &74.12 &28.50 &2.50 &31.81 &43.65 &41.19 &27.83 \\
&H2O &21.66 &21.64 &38.60 &30.96 &20.63 &13.02 &20.65 &22.61 &22.08 &39.00 &82.19 &39.75 &3.16 &79.98 &51.25 &48.20 &34.71 \\
&SnapKV &20.11 &21.28 &42.98 &37.51 &22.31 &14.43 &19.19 &21.89 &21.01 &48.00 &83.77 &40.44 &2.51 &66.99 &51.64 &48.57 &35.16 \\
&PyramidKV &22.11 &22.52 &43.04 &33.57 &22.98 &15.69 &20.56 &22.52 &21.36 &65.50 &83.84 &40.03 &2.89 &67.26 &51.51 &46.42 &36.36 \\
&ours &22.05 &23.65 &43.08 &36.03 &22.60 &15.23 &21.35 &23.11 &22.19 &68.00 &84.79 &41.02 &4.20 &70.11 &52.45 &47.41 &\textbf{37.33} \\

\midrule

\multirow{6}{*}{\rotatebox{90}{\textbf{\makecell{Qwen2-7B\\-Instruct}}}}
 &FullKV &25.14 &42.35 &45.04 &14.80 &14.13 &9.23 &36.35 &23.79 &26.51 &76.50 &89.16 &45.23 &6.50 &75.50 &60.30 &60.78 &40.71 \\
\cmidrule(lr){2-19}
&StreamingLLM &19.25 &23.63 &26.51 &14.00 &15.30 &7.46 &18.07 &19.30 &18.30 &47.00 &77.92 &31.57 &6.50 &17.00 &42.52 &41.94 &26.64 \\
&H2O &20.33 &30.43 &34.22 &13.61 &13.37 &7.81 &20.72 &21.66 &18.44 &40.00 &86.94 &42.17 &7.00 &70.50 &53.45 &53.76 &33.40 \\
&SnapKV &22.26 &31.62 &38.95 &16.05 &17.71 &7.66 &18.91 &21.41 &18.21 &46.00 &87.61 &42.01 &6.50 &63.50 &54.87 &53.03 &34.14 \\
&PyramidKV &20.50 &31.70 &39.95 &18.54 &18.54 &8.85 &19.24 &20.47 &18.18 &60.00 &87.98 &39.71 &7.00 &49.00 &48.77 &47.91 &33.52 \\
&ours &22.77 &35.57 &42.62 &14.80 &16.35 &8.31 &21.41 &21.97 &19.56 &58.00 &88.18 &40.93 &6.50 &70.00 &53.58 &52.50 &\textbf{35.82} \\

\midrule

\multirow{6}{*}{\rotatebox{90}{\textbf{\makecell{InternLM-2.5-7B\\-Chat-1M}}}}
 &FullKV &22.42 &27.61 &39.98 &40.92 &33.48 &26.68 &33.01 &25.18 &26.28 &72.50 &86.76 &39.76 &2.91 &100.00 &55.86 &57.95 &43.21 \\
\cmidrule(lr){2-19}
&StreamingLLM &17.91 &13.02 &24.31 &24.27 &16.01 &11.29 &17.29 &20.62 &18.06 &48.5 &67.53 &21.93 &0.82 &87.39 &43.45 &42.79 &29.70 \\
&H2O &16.16 &17.71 &27.94 &26.83 &17.83 &17.81 &13.99 &22.59 &16.9 &39.50 &81.87 &32.15 &1.32 &96.50 &48.30 &47.27 &32.79 \\
&SnapKV &19.65 &17.44 &35.29 &27.36 &18.58 &19.79 &12.76 &22.42 &16.31 &48.00 &80.23 &31.35 &0.95 &95.00 &49.47 &48.22 &33.93 \\
&PyramidKV &18.80 &17.35 &33.48 &31.16 &20.05 &19.02 &14.65 &22.02 &17.40 &69.50 &80.87 &32.02 &1.23 &95.00 &47.13 &44.73 &35.28 \\
&ours &17.93 &19.89 &34.15 &31.50 &19.03 &20.60 &15.14 &22.41 &18.15 &70.00 &83.09 &32.44 &0.86 &95.50 &49.33 &47.16 &\textbf{36.07} \\

\bottomrule
\end{tabular}
}
\caption{\textbf{Performance comparison on the LongBench dataset} for full KV cache, previous methods (StreamingLLM, H2O, SnapKV, PyramidKV), and our DynamicKV method, with KV cache sizes of 128, using models including LLaMA3-8B-Instruct, Mistral-7B-Instruct-v0.2, QWen2-7B-Instruct, and InternLM-2.5-Chat-1M. Bold indicates the best performance.}
\label{table:longbench_128}
\vspace{-3mm}
\end{table*}

Table~\ref{table:longbench_128} presents a performance comparison on the LongBench for different KV cache compression methods (StreamingLLM, H2O, SnapKV, PyramidKV and our DynamicKV) with a fixed cache size of 128. We conduct experiments across various tasks such as \itbf{Single-Document QA, Multi-Document QA, Summarization, Few-shot Learning, Synthetic tasks}, and \itbf{Code Completion.} 

The results show that our DynamicKV consistently achieves competitive or superior performance compared to previous methods. While FullKV yields the highest average scores, DynamicKV achieves the best or near-best performance across several models -- particularly excelling with Mistral-7B-Instruct-v0.2 and InternLM-2.5-Chat-1M -- demonstrating effective memory compression with minimal loss in accuracy.

\end{document}